\documentclass[runningheads]{llncs}
\usepackage{hyperref}       
\usepackage{url}            
\usepackage{booktabs}       
\usepackage{amsfonts}       
\usepackage{nicefrac}       
\usepackage{microtype}      
\usepackage[pdftex]{graphicx}
\usepackage{enumitem}
\usepackage{amsmath}
\begin{document}
\title{Hierarchical Bayesian Regression for Multi-Site Normative Modeling of Neuroimaging Data}
\titlerunning{Hierarchical Bayesian Regression for Multi-Site Normative Modeling}
%
\author{Seyed Mostafa Kia\inst{1,2} \and
Hester Huijsdens\inst{1} \and
Richard Dinga\inst{1,2} \and
Thomas Wolfers\inst{1,3,4} \and
Maarten Mennes\inst{1} \and
Ole A. Andreassen\inst{3,4} \and
Lars T. Westlye\inst{3,4} \and
Christian F. Beckmann\inst{1,2} \and
Andre F. Marquand\inst{1,2,5}}
\authorrunning{S.M. Kia et al.}
%
\institute{Donders Institute, Radboud University, Nijmegen, the Netherlands \and
Radboud University Medical Centre, Nijmegen, the Netherlands \and
Department of Psychology, University of Oslo, Norway \and
NORMENT, University of Oslo and Oslo University Hospital, Oslo, Norway \and 
Institute of Psychiatry, King’s College London, London, UK \\
}
\maketitle              
\setcounter{footnote}{0} 
\begin{abstract}
Clinical neuroimaging has recently witnessed explosive growth in data availability which brings studying heterogeneity in clinical cohorts to the spotlight. Normative modeling is an emerging statistical tool for achieving this objective. However, its application remains technically challenging due to difficulties in properly dealing with nuisance variation, for example due to variability in image acquisition devices. Here, in a fully probabilistic framework, we propose an application of hierarchical Bayesian regression (HBR) for multi-site normative modeling. Our experimental results confirm the superiority of HBR in deriving more accurate normative ranges on large multi-site neuroimaging data compared to widely used methods. This provides the possibility i) to learn the normative range of structural and functional brain measures on large multi-site data; ii) to recalibrate and reuse the learned model on local small data; therefore, HBR closes the technical loop for applying normative modeling as a medical tool for the diagnosis and prognosis of mental disorders. 
\end{abstract}
\section{Introduction}      
\label{sec:introduction}
Neuroimaging has recently entered the era of big data. This has ignited a movement in the clinical neuroimaging community toward understanding the heterogeneous neurobiological underpinnings of mental disorders in large and demographically diverse populations~\cite{thompson2014enigma}. Achieving this goal in practice requires aggregating neuroimaging datasets usually acquired at several imaging centers with different acquisition protocols and scanners. This diversity in acquisition protocols and associated site-related peculiarities introduces significant site-related variability in the data~\cite{fortin2018harmonization}, referred to as \emph{site-effects}, that severely confound the result of any subsequent analyses.

\emph{Normative modeling} is recently introduced as a statistical tool for studying the biological variability of mental disorders in clinical neuroimaging cohorts~\cite{marquand2016understanding}. The first step in normative modeling is to estimate the centiles of variation, \textit{i.e.,} the normative ranges, of the biological measures as a function of clinical covariates. This is performed via regressing the units of neuroimaging data (\textit{e.g.}, a voxel in structural or functional MRIs) against a set of clinically relevant covariates (\textit{e.g.}, demographics). The deviations of clinical samples from the resulting normative range can be quantified as Z-scores at the individual level~\cite{marquand2019conceptualizing}. Such deviations can be interpreted as individualized biomarkers for psychiatric disorders~\cite{wolfers2018mapping,zabihi2019dissecting} in the spirit of \emph{precision psychiatry}. However, normative modeling on multi-site data is challenging because combining several neuroimaging datasets introduces artefactual variability that confounds the derived deviations~\cite{marquand2019conceptualizing}. This limits the practical application of normative modeling as a medical tool because the data collected at different centers may have different characteristics. Thus, it is essential to develop \emph{adaptive} methods that can effectively deal with site-effects.

The most prevalent approach to deal with site-effects is to regress them out from the data. A popular method is ComBat~\cite{johnson2007adjusting} that has been adopted from genomics for harmonizing neuroimaging data. ComBat uses an empirical Bayes method for adjusting additive and multiplicative batch effects in data. It has shown great potential in harmonizing different neuroimaging data modalities including diffusion tensor imaging~\cite{fortin2017harmonization}, cortical thickness~\cite{fortin2018harmonization}, and structural/functional MRIs~\cite{nielson2018detecting,yamashita2019harmonization,pompinio2020harmonization}. However, this approach comes with two limitations. From a theoretical point of view, ComBat regresses all variance associated with site and only preserves \textit{a priori} known sources of variation in data, which are accounted for in the design matrix. In other words, it is necessary to specify in advance which shared variation should be retained. This can be restrictive especially when we are interested in exploratory analysis of unknown biological factors. An illustrative example is stratifying psychiatric disorders into subtypes~\cite{marquand2016beyond}. Since subtypes are unknown in advance, their biological correlates in brain images can be removed or corrupted. Moreover, in many cases, clinical covariates (such as age) strongly correlate with site-effects, thus, any effort toward removing site-effects may result in losing the signal of interest. From a practical perspective, it is difficult to apply current implementations of ComBat to data coming from new sites since ComBat requires access to data from all sites at training time to compute shared parameters including intercept, regression coefficients, and noise variance. This obstacle is even more pronounced when dealing with large cohorts where it may not be possible to share the data, \textit{e.g.,} due to data anonymity concerns or a lack of ethical permission for data sharing~\cite{poline2010data}.

To overcome these limitations, we propose hierarchical Bayesian regression (HBR)~\cite{gelman2013bayesian} for probabilistic modeling of batch-effects in neuroimaging data. In this framework, we impose shared prior distributions over site-specific model parameters. Our method has several appealing features: i) it is fully probabilistic, thus, it is well-suited to normative modeling as it provides estimations of both phenomenological variability in data and epistemological uncertainty in the model~\cite{cox2006principles}; ii) it preserves all sources of variation in the data, which overcomes the requirement to specify in advance which parts of the variance will be retained; iii) it is highly flexible and accommodates different modeling choices (\textit{e.g.}, non-linear effects or heteroscedastic noise); iv) it provides the possibility of transferring  hyperpriors of a reference model when recalibrating the normative model to data from new sites. Using a large dataset of 7499 participants aggregated across 33 scanners, we show the potential of HBR in estimating the predictive posterior distribution compared to ComBat and trivial pooling. We also demonstrate an application of the proposed framework in understanding the biological signatures of several brain disorders.

\section{Methods}      
\label{sec:methods}
\subsection{Normative Modeling}
\label{subsec:normative_modeling}
Let $\mathbf{X} \in \mathbb{R}^{n \times p}$ represent a matrix of $p$ clinical covariates for $n$ participants. We denote the corresponding neuroimaging measures at each measurement unit (\textit{e.g.}, a voxel) by $\mathbf{y} \in \mathbb{R}^n$. Assuming a Gaussian distribution over each neuroimaging measure, \textit{i.e.,} $y \sim \mathcal{N}(\mu,\sigma^2)$, in normative modeling we are interested in finding a parametric or non-parametric form for $\mu$ and $\sigma$ given the covariates $\mathbf{X}$.\footnote{Here, for generality we specify heteroscedastic noise to model age-dependent variance. The homoscedastic formulation is a special case where $\sigma$ is independent of $\mathbf{X}$.} This is achieved by estimating $f_\mu(\mathbf{X}, \theta_\mu)$ and $f_\sigma^+(\mathbf{X}, \theta_\sigma)$, where $f^+$ is a non-negative function;\footnote{Non-negativity can be enforced for example using a softplus function $f_\sigma^+=log(1+f_\sigma)$.} $\theta_\mu$ and $\theta_\sigma$ are respectively the parameters of $f_\mu$ and $f_\sigma^+$. Then, for example, $\mu \pm 1.96 \sigma$ forms the $95 \%$ percentile for the normative range of $\mathbf{y}$. The deviations of samples from the normative range is quantified as Z-scores~\cite{marquand2016understanding}:
\begin{eqnarray} \label{eq:deviation}
\mathbf{z} = \frac{\mathbf{y} - f_\mu(\mathbf{X}, \theta_\mu)}{f_\sigma^+(\mathbf{X}, \theta_\sigma)}.
\end{eqnarray}

Any large deviation from the normative range is interpreted as an abnormality in the brain's structure or function and can be studied concerning different mental disorders. The abnormal probability index for each sample can be computed by translating the deviations (Z-scores) to the corresponding p-values.

\subsection{Problem Statement: Multi-Site Normative Modeling}
\label{subsec:problem}
\begin{figure}[t!]
	\centering
	\includegraphics[width=0.95\textwidth]{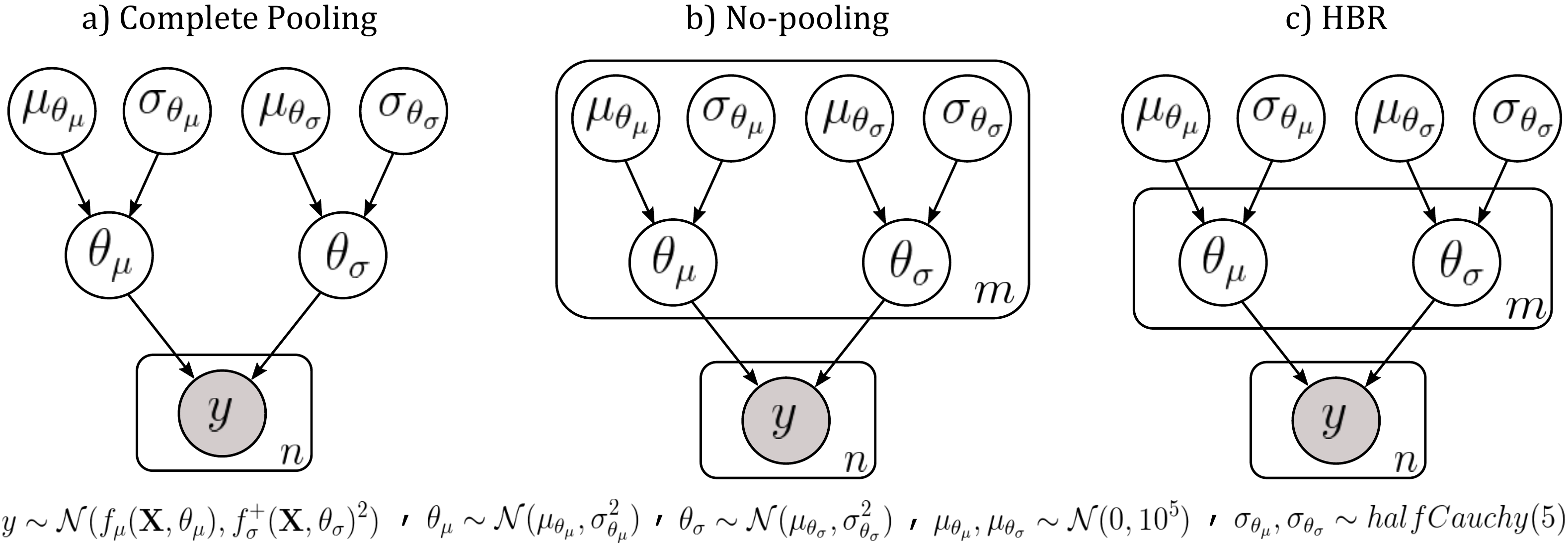}
	\caption{Graphical models of a) pooling, b) no-pooling, and c) partial-pooling via HBR.}
	\label{fig:graphical_models}
	\vspace{-0.5cm}
\end{figure}

Let $\mathbf{y}_{i} \in \mathbb{R}^{n_i}$ denote neuroimaging measures for $n_i$ participants in the $i$th batch, $i \in \{1,\dots,m\}$, of data and we have $y_i \sim \mathcal{N}(\mu_i,\sigma_i^2)$. Here, each batch refers to data which are collected at different imaging sites, however, our formulations are general for other possible batch-effects in data (\emph{e.g.}, gender). There are three possible strategies for normative modeling on multi-batch data:
\begin{enumerate}[leftmargin=*]
\item \textbf{Complete Pooling:} where the batch-effects in data are ignored by assuming $y_1,\dots,y_m \sim \mathcal{N}(\mu,\sigma^2)$ and we have:
\begin{eqnarray} \label{eq:pooling}
\mathbf{y}_i = f_\mu(\mathbf{X}, \theta_\mu) + \epsilon \quad \forall i \in \{1,\dots,m\},
\end{eqnarray}

where $\epsilon$ is zero-mean error with standard deviation $f_\sigma^+(\mathbf{X}, \theta_\sigma)$. In complete pooling, parameters ($\theta_\mu$ and $\theta_\sigma$) and hyperparameters ($\mu_{\theta_\mu}$, $\sigma_{\theta_\mu}$, $\mu_{\theta_\sigma}$, and $\sigma_{\theta_\sigma}$) of the mean and variance are fixed across batches (Fig.~\ref{fig:graphical_models}a). Even though the pooling approach provides a simple solution to benefit from a larger sample size, the assumption that data from different batches have identical distributions is very limiting and restricts its usage in normative modeling because batch-effects will be encoded in the resulting deviations in Eq.~\ref{eq:deviation}.

\item \textbf{Harmonization:} that can be considered as a corrected pooling scenario in which we try to adjust the location and scale of data density for batch-effects. ComBat~\cite{johnson2007adjusting} is a common and effective harmonization technique for neuroimaging data where we have:
\begin{eqnarray} \label{eq:harmonization}
\tilde{\mathbf{y}}_i = \frac{\mathbf{y}_i - g(\mathbf{X}) - \gamma_i}{\delta_i} + g(\mathbf{X}),
\end{eqnarray}
where $\tilde{\mathbf{y}}_i$ is harmonized data that is expected to be homogeneous across batches; $\gamma_i$ and $\delta_i$ are respectively the additive and multiplicative batch-effects. Here, $g(\mathbf{X})$ is a linear or non-linear~\cite{poline2010data} function that preserves the signal of interest as specified in the design matrix $\mathbf{X}$. After harmonization, Eq.~\ref{eq:pooling} can be used for regressing the data. Using ComBat for multi-site normative modeling comes with the limitations described above (\textit{i.e.}, it only preserves the known sources of variation in the data specified in $\mathbf{X}$ and all data should be available when estimating the parameters of $g(\mathbf{X})$). 

\item \textbf{No-pooling:} in which separate models are estimated for each batch (Fig.~\ref{fig:graphical_models}b):
\begin{eqnarray} \label{eq:unpooling}
\mathbf{y}_i = f_{\mu_i}(\mathbf{X}, \theta_{\mu_i}) + \epsilon_i \quad i \in \{1,\dots,m\}.
\end{eqnarray}
No-pooling is immune to problems of complete pooling and harmonization, however, it cannot take full advantage of the large sample size. It is also prone to overfitting especially when $f_{\mu_i}$ and $f_{\sigma_i}^+$ are complex functions and the number of samples in each batch is small. This may result in spurious and inconsistent estimations of parameters of the model across different batches.
\end{enumerate}  

Considering the shortcomings of aforementioned methods, there is an emergent need to find an alternative approach that i) similar to complete pooling and harmonization methods benefits from advantages of big data; ii) is immune to batch-effects in the resulting deviations, likewise ComBat and no-pooling; iii) unlike ComBat preserves unknown sources of biological variation in data; iv) provides the possibility to reapply the normative model to data from new sites. 

\subsection{Partial-Pooling using Hierarchical Bayesian Regression}
\label{subsec:solution}
Hierarchical Bayesian regression (HBR) is a natural choice in modeling different levels of variation in data~\cite{gelman2013bayesian}. In HBR, the structural dependencies between variables are incorporated in the modeling process by coupling them via a shared prior distribution. To adopt HBR for multi-site normative modeling, we assume $\theta_{\mu_i}$ and $\theta_{\sigma_i}$ in Eq.~\ref{eq:unpooling} (that govern the data generating process for each batch $\mathbf{y}_i$) are coming exchangeably from the same prior distribution, \textit{i.e.}, $\forall i, \theta_{\mu_i} \sim \mathcal{N}(\mu_{\theta_\mu},\sigma_{\theta_\mu}^2)$ and $\theta_{\sigma_i} \sim \mathcal{N}(\mu_{\theta_\sigma},\sigma_{\theta_\sigma}^2)$ (see Fig.~\ref{fig:graphical_models}c). Such a joint prior acts like a regularizer over parameters of the model and prevents it from the overfitting on small batches. In fact, HBR allows for a reasonable compromise between the complete pooling and no-pooling scenarios as it combines all models in Eq.~\ref{eq:unpooling} into a single model that benefits from the wealth of big data.

Importantly, HBR also provides the possibility to transfer the knowledge inferred about the distribution of hyperparameters from a primary set of observed data $\mathbf{y}$ to secondary datasets from new sites $\mathbf{y}^*$. To achieve this, we propose to use posterior distributions of hyperparameters, \textit{i.e.}, $p(\mu_{\theta_\mu} \mid \mathbf{y})$, $p(\sigma_{\theta_\mu} \mid \mathbf{y})$, $p(\mu_{\theta_\sigma} \mid \mathbf{y})$, and $p(\sigma_{\theta_\sigma} \mid \mathbf{y})$, as \emph{informative} hyperpriors for the secondary model. Informative hyperpriors enable us to incorporate pre-existing evidence when re-inferring the model on new data rather than ignoring it by non-informative or weakly informative hyperpriors. This is a critical feature for model portability because it enables effective model recalibration without the need to having access to the primary set of data.

\begin{table}[!b]
\centering
\caption{Demographics of multi-site experimental data.}
\label{tab:datasets}
\resizebox{0.8\textwidth}{!}{%
\begin{tabular}{@{}ccccccc@{}}
\toprule
\textbf{Datasets} &
  \textbf{\begin{tabular}[c]{@{}c@{}}No. \\ Scans\end{tabular}} &
  \textbf{\begin{tabular}[c]{@{}c@{}}No. \\ Patients\end{tabular}} &
  \textbf{\begin{tabular}[c]{@{}c@{}}No. \\ Scanners\end{tabular}} &
  \textbf{\begin{tabular}[c]{@{}c@{}}Age \\ Range\end{tabular}} &
  \textbf{\begin{tabular}[c]{@{}c@{}}Gender\\ M/F\end{tabular}} &
  \textbf{\begin{tabular}[c]{@{}c@{}}FS\\ Version\end{tabular}} \\ \midrule
FCON1000~\cite{biswal2010toward}         & 1094 & 25(ADHD)                            & 22 & 8--85  & 494/600   & 6.0 \\
CAMCAN~\cite{taylor2017cambridge}        & 647  & -                                   & 1  & 18--88 & 318/329   & 6.0 \\
PNC~\cite{satterthwaite2016philadelphia} & 1514 & -                                   & 1  & 8--23  & 731/783   & 6.0 \\
HCP1200~\cite{vanessen2010human}         & 1113 & -                                   & 1  & 22--37 & 507/606   & 5.3 \\
OASIS3~\cite{lamontagne2019oasis}        & 2044 & 271(AZ),51(MCI)                     & 5  & 43--97 & 866/1178  & 5.3 \\
TOP~\cite{skaatun2016global}             & 823  & 167(SZ),193(BD),31(MDD),107(others) & 1  & 17--69 & 435/388   & 6.0 \\
CNP~\cite{poldrack2016phenome}           & 264  & 49(SZ),49(BD),41(ADHD)              & 2  & 21--50 & 152/112   & 6.0 \\ \midrule
\textbf{Total}                           & 7499 & 1017                                & 33 & 8--97  & 3503/3996 & -   \\ \bottomrule
\end{tabular}%
}
\end{table}

\section{Experiments and Results}      
\label{sec:experiments}
\subsection{Experimental Materials}      
\label{subsec:materials}
\begin{figure}[t!]
	\centering
	\includegraphics[width=0.9\textwidth]{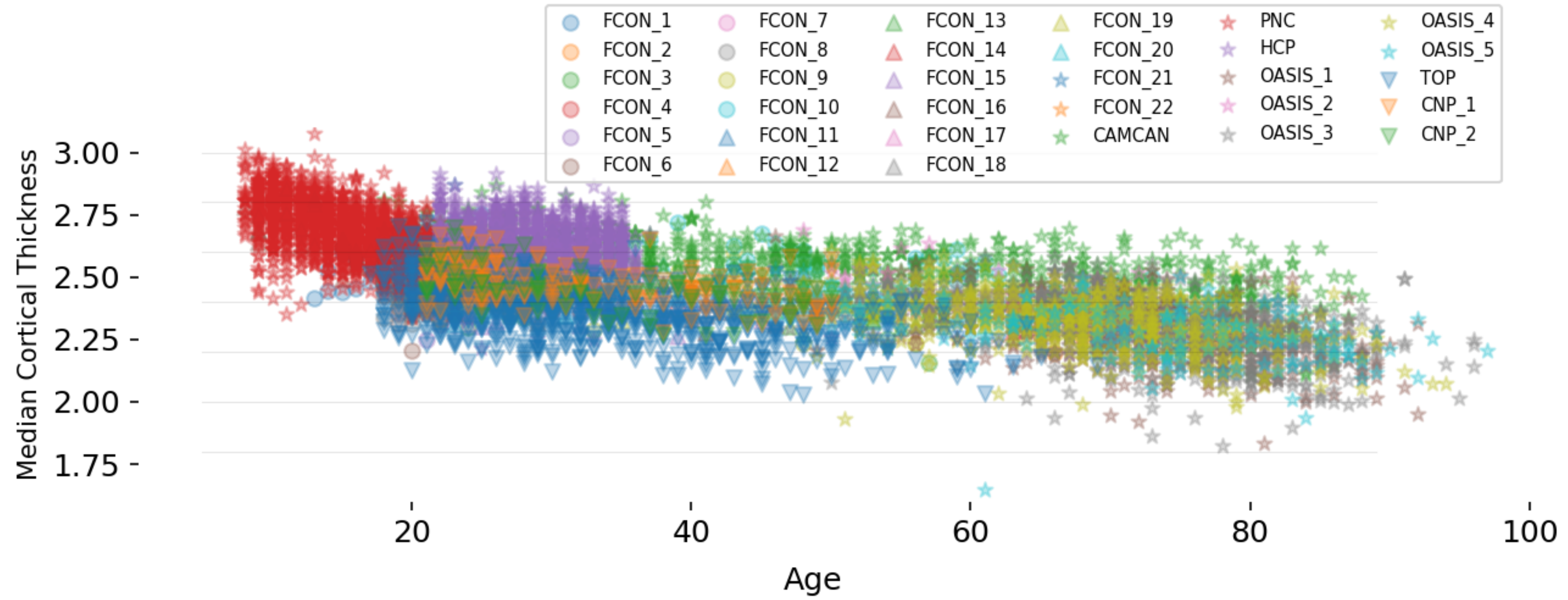}
	\caption{Site-effect in the distribution of median cortical thicknesses across 33 scanners.}
	\label{fig:data}
	\vspace{-0.5cm}
\end{figure}

Table~\ref{tab:datasets} lists the 7 neuroimaging datasets that are used in our experiments. Low-quality scans and participants with missing demographic information are excluded. The final data consist of 7499 scans from 7 datasets including 33 scanners that reasonably cover a wide range of human lifespan from 8 to 97 years old (see supplement for the age distribution). These properties make these data a perfect case-study for large-scale multi-site normative modeling of aging. The data also contain 1017 scans from participants diagnosed with a neurodevelopmental, psychiatric, or neurodegenerative disease including attention deficit hyperactivity disorder (ADHD), schizophrenia (SZ), bipolar disorder (BD), major depressive disorder (MDD), mild cognitive impairment (MCI), and Alzheimer's disease (AZ). In our analyses, we use cortical thickness measures estimated by Freesurfer~\cite{fischl2012freesurfer} over 148 cortical regions in the Destrieux atlas~\cite{destrieux2010automatic}. Fig.~\ref{fig:data} shows the distribution of median cortical thickness across participants and scanners.  

\subsection{Implementations and Model Settings}      
\label{subsec:implementations}
All methods are implemented using the PyMC3 package~\cite{salvatier2016probabilistic}. A No-U-Turn sampler (NUTS)~\cite{hoffman2014no} is used for inferring the posterior distributions of parameters and hyperparameters. Given the characteristics of experimental data, in all experiments, we opt to use a linear form for $f_\mu$ and a homoscedastic form for the variance.\footnote{We emphasize that the proposed framework is capable of modeling heteroscedasticity. Here, using a heteroscedastic model for the variance did not provide any advantage at the cost of higher model complexity (see Fig.~S\ref{fig:regression_het} in the supplement).} Normal and half-Cauchy distributions are respectively used as hyperpriors for the mean and standard deviation of parameters of $f_\mu$ (see Fig.~\ref{fig:graphical_models}). The distribution of the standard deviation of the homoscedastic noise is set to a uniform distribution in the range of 0 to 100. Non-centered parameterizations are used to simplify posterior geometries and increase the performance of the sampler~\cite{betancourt2015hamiltonian}. For harmonizing data using ComBat, we use a Python implementation available at \url{https://github.com/Warvito/neurocombat_sklearn}. All implementations are available online at~\url{https://github.com/amarquand/nispat}.

\subsection{Experimental Setup and Results}      
\label{subsec:results}
We set up two experimental settings, regression and anomaly detection. In all experimental configurations, we use age as a covariate (in $\mathbf{X}$ with $p=1$) and gender is dealt with as a batch-effect. In the HBR case, the site is also included as a batch-effect. All experiments and evaluations are repeated 10 times with different random healthy participants in the training and test phases. 

In the regression setting, the goal is to compare the accuracy of HBR with its alternatives in deriving the normative range of cortical thickness in a healthy population across the human lifespan. Here, we assume the data from all scanners are available when estimating the normative model. In each experimental run, $80\%$ of healthy samples are randomly selected to train the regression model and the remaining $20\%$ are used for the evaluation. We use three metrics to evaluate the resulting normative models, i) Pearson's correlation coefficient (RHO); ii) standardized mean squared error (SMSE); iii) mean standardized log-loss (MSLL). While correlation and SMSE evaluate only the predicted mean, MSLL also accounts for the quality of estimated variance which plays an important role in deriving deviations from the norm (see Eq.~\ref{eq:deviation}).

\begin{figure}[t!]
	\centering
	\includegraphics[width=0.975\textwidth]{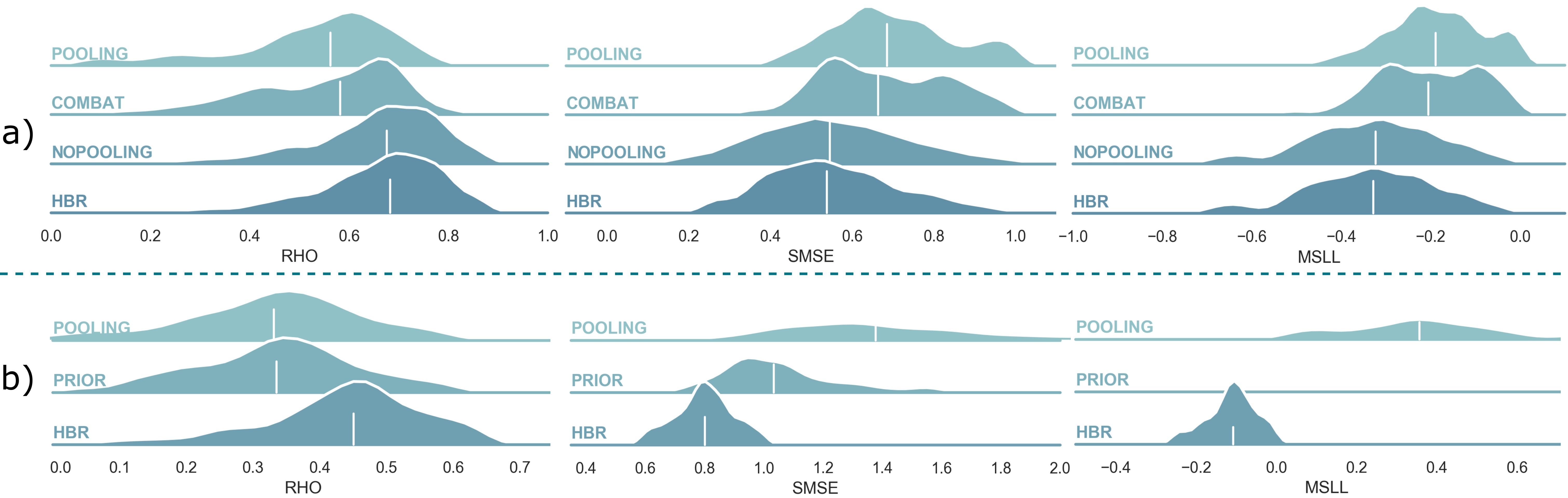}
	\caption{The distributions of correlation, SMSE, and MSLL across 148 cortical areas and 10 experimental repetitions in the a) regression, and b) anomaly detection settings. The white lines highlight the medians of distributions.}
	\label{fig:regression}
	\vspace{-0.5cm}
\end{figure}

Fig.~\ref{fig:regression}a compares the densities of our evaluative metrics (across 148 cortical areas and 10 runs) in the regression scenario. In all cases, HBR and no-pooling show better performance compared to pooling and ComBat harmonization. This boost in regression performance is gained by accounting for the difference between the distributions of signal and noise across different sites, rather than ignoring or removing it. Considering the simplicity of the employed linear parameterization for modeling the mean (thus less chance for overfitting), the improvement of HBR in comparison to no-pooling remains negligible. Furthermore, to ensure that the resulting deviations are not contaminated with residual site bias, we used a linear support vector machine on derived deviations to classify scanners in a one-versus-all setting. Balanced classification accuracy was at chance-level for the HBR, harmonization and no-pooling whereas under the pooling condition scanners were discriminated with $71\%$ accuracy, indicating clear site-effects.

In the anomaly detection setting, we demonstrate an application of HBR in a more realistic clinical scenario when applying a reference normative model on new data from new scanners (thus ComBat cannot be applied in this scenario). To do so, we first estimate the parameters of the reference normative model on FCON1000, CAMCAN, PNC, and HCP datasets. Then, in each run $50\%$ of random healthy participants in OASIS3, TOP, and CNP datasets are used to recalibrate the model. The rest of the healthy participants and patients are used as test samples. Fig.~\ref{fig:regression}b compares the regression performance of HBR with pooling and only HBR priors (where we roughly set $\forall i, \theta_{\mu_i}=\mu_{\theta_\mu}, \theta_{\sigma_i}=\mu_{\theta_\sigma}$). HBR performs best in predicting data from new sites. Using only priors of HBR for prediction on new sites provides a reasonable estimate of the predictive mean, but extremely poor estimates of the predictive variance (its MSLL density lies fully outside of the plotted range).

\begin{figure}[t!]
	\centering
	\includegraphics[width=0.75\textwidth]{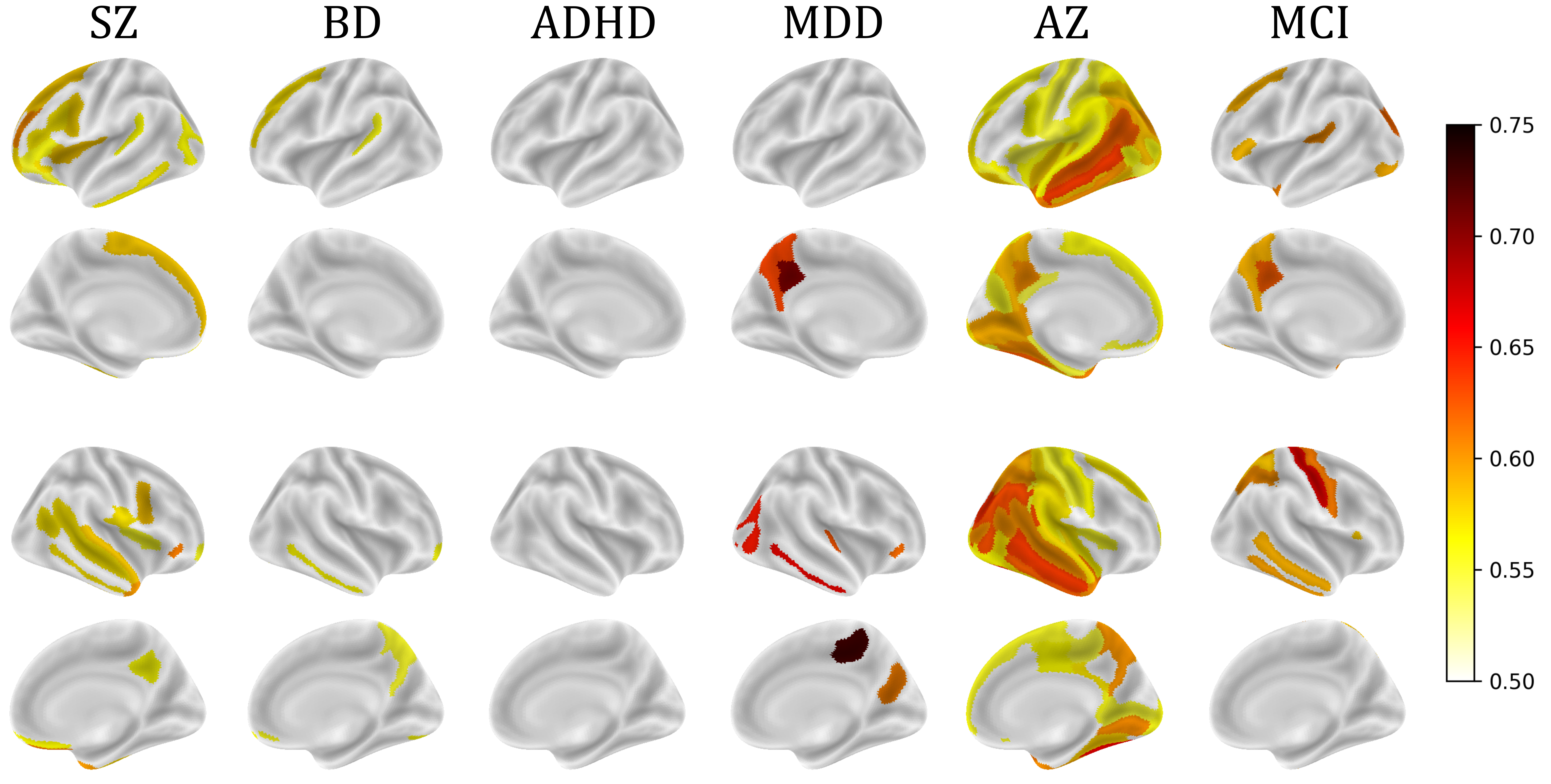}
	\caption{AUCs across brain regions in detecting healthy participants from patients.}
	\label{fig:AUCs}
	\vspace{-0.5cm}
\end{figure}

We further compute the abnormal probability index for each individual across 148 cortical regions (see Sec.~\ref{subsec:normative_modeling}), and use the area under the ROC curve (AUC) to evaluate the predictive power of deviations for each diagnosis. Fig.~\ref{fig:AUCs} depicts the resulting significant and stable AUCs across brain regions. To test for the significance, we performed permutation tests with 1000 repetitions and used $0.05$ as the threshold. To ensure stability, only significant areas that are stable across 10 repetitions are kept. Except for ADHD, the resulting significant detection performances show that the deviation from the normative range of cortical thickness contains valuable information regarding the brain's structural changes in different disorders. The spatial distribution of discriminative regions largely overlaps with brain areas that are known to be implicated in the corresponding disorders. For example, patients with SZ have lower cortical thickness compared to the healthy population in the right medial orbital sulcus, right orbital inferior frontal gyrus, and left middle frontal sulcus (see supplementary Fig.~S\ref{fig:scatter_plots} for fits).  

\section{Summary}      
\label{sec:summary}
Here, we introduced a fully probabilistic framework that accommodates signal and noise variance in multi-site neuroimaging data via estimating different but connected mean and variance components across different sites. This is a key feature relative to the usual approach of regressing out site effects, especially when the scientific question lies in understanding heterogeneity in large cohorts. The proposed framework is quite general and accommodates many different parametric/non-parametric and linear/non-linear forms for modeling the signal mean and homoscedastic/heteroscedastic variance. Further, it provides the possibility to construct a universal normative model on massive data samples and, after recalibration, reuse it on local data for prediction of brain disorders.

\section*{Acknowledgements}      
\label{sec:acknowledgements}
This work was supported by the Dutch Organisation for Scientific Research via Vernieuwingsimpuls VIDI fellowships to AM (016.156.415) and CB (864.12.003). The authors also gratefully acknowledge support from the Wellcome Trust via digital Innovator (215698/Z/19/Z) and strategic awards (098369/Z/12/Z).

\bibliographystyle{splncs04}
\bibliography{references}

\newpage
\renewcommand{\figurename}{Fig. S}
\setcounter{figure}{0}    

\section*{Supplementary Materials}
\label{sec:supplementary}

\begin{figure}[h!]
	\centering
	\includegraphics[width=0.5\textwidth]{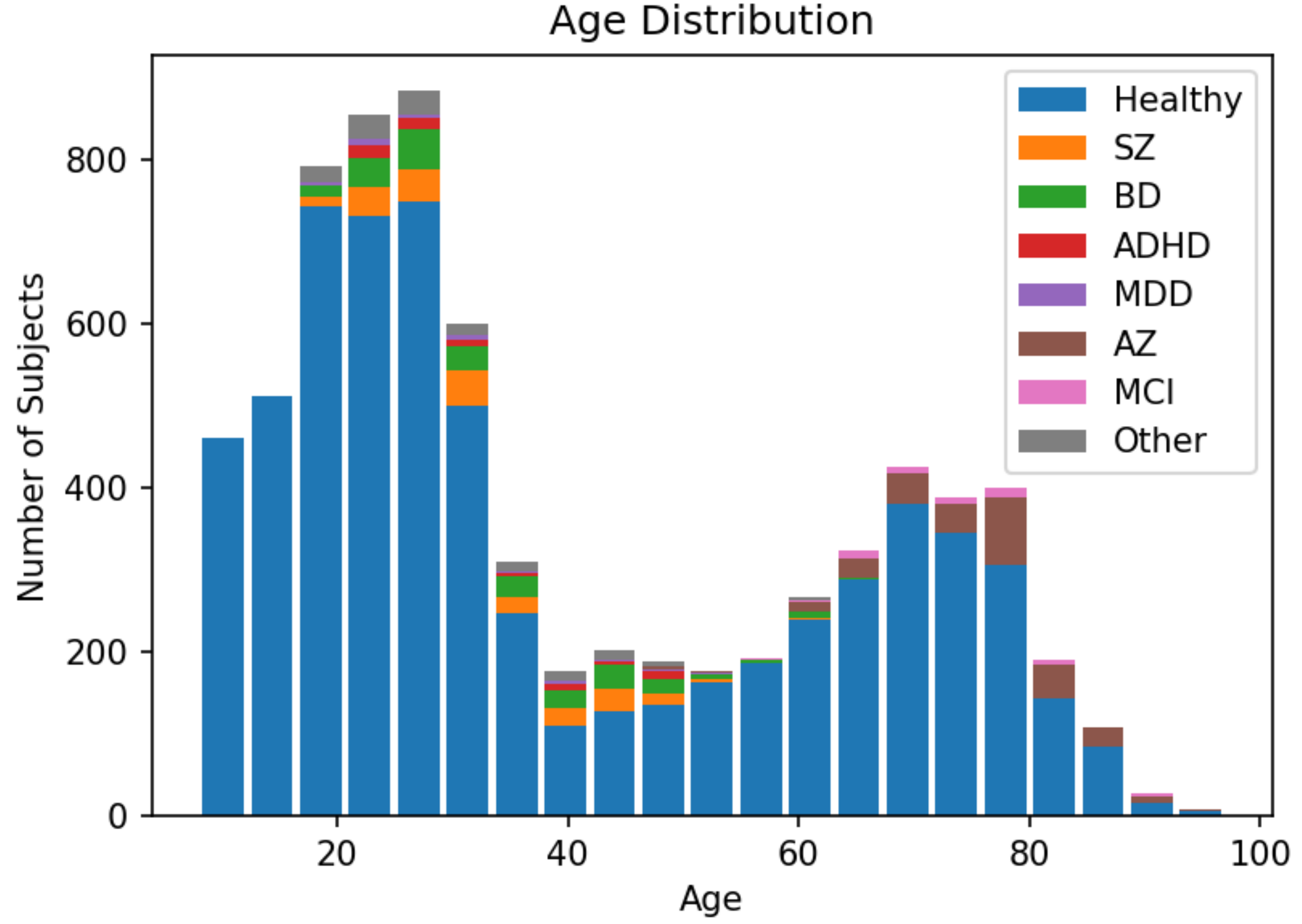}
	\caption{Histogram of age and diagnosis distributions in the aggregated dataset.}
	\label{fig:age_distribution}
	\vspace{-0.5cm}
\end{figure}

\begin{figure}[h!]
	\centering
	\includegraphics[width=0.95\textwidth]{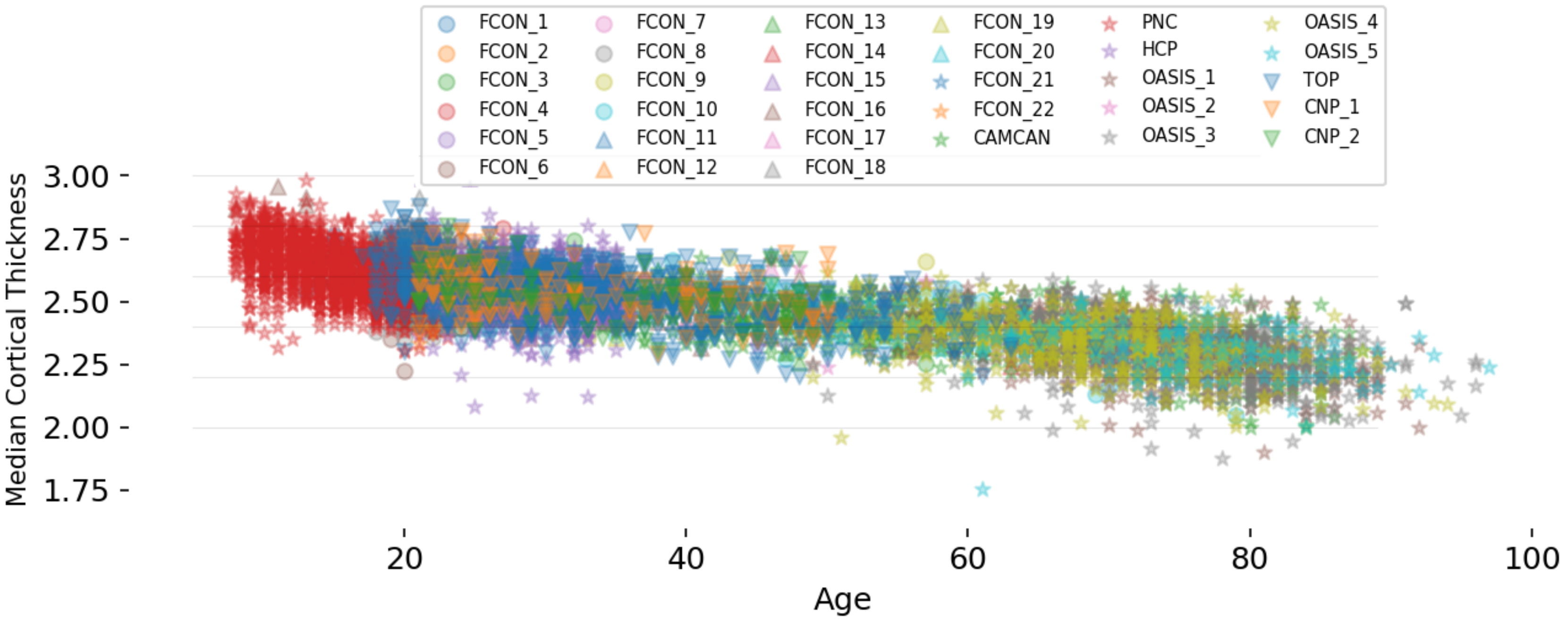}
	\caption{Median cortical thicknesses across 33 scanners after ComBat harmonization.}
	\label{fig:harmonized_data}
	\vspace{-0.5cm}
\end{figure}

\begin{figure}[h!]
	\centering
	\includegraphics[width=0.9\textwidth]{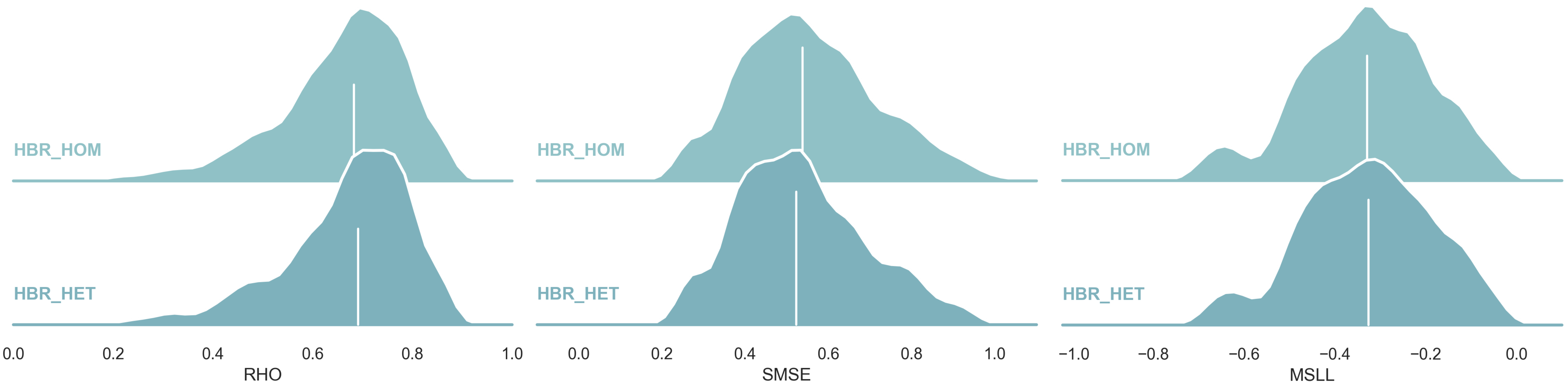}
	\caption{Comparison between regression performance of HBRs with homoscedastic and heteroscedastic variance. In the heteroscedastic case, a quadratic polynomial model is used for modeling the variance. Given the characteristics of experimental data, using a heteroscedastic model for the variance does not improve the regression performance.}
	\label{fig:regression_het}
\end{figure}

\begin{figure}[h!]
	\centering
	\includegraphics[width=0.995\textwidth]{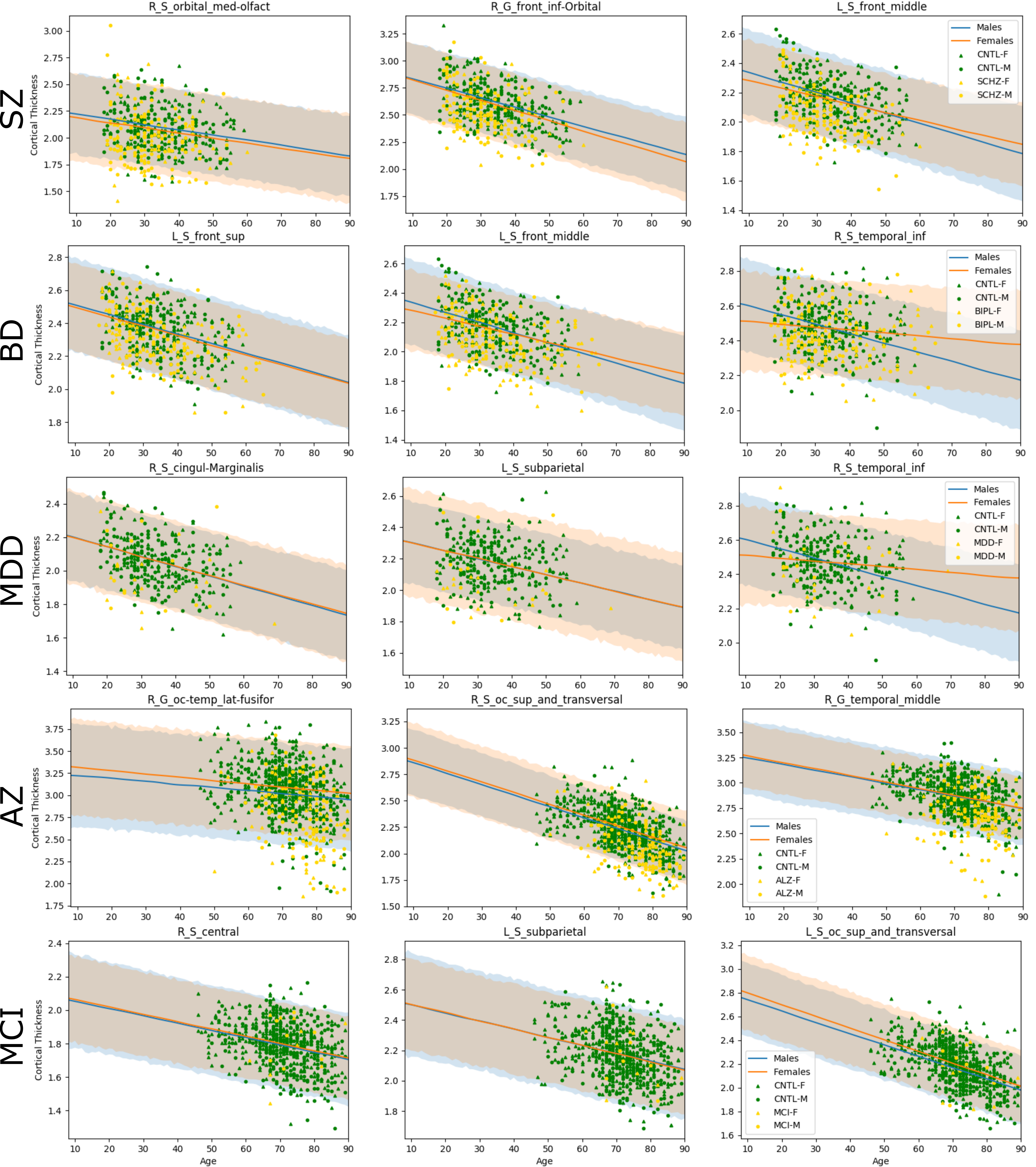}
	\caption{Example HBR normative ranges of cortical thicknesses for the 3 most discriminative regions (columns) and 5 diagnoses (rows). For each diagnosis, only the data from one scanner is plotted.}
	\label{fig:scatter_plots}
\end{figure}

\end{document}